\documentclass[final]{cvpr}

\usepackage{amsmath}
\usepackage{amsfonts}
\usepackage{dsfont}

\usepackage[T1]{fontenc}
\usepackage[utf8]{inputenc}
\usepackage{booktabs}
\usepackage{dcolumn}
\usepackage{units}
\usepackage{array}
\usepackage{graphicx}
\usepackage{graphics}
\usepackage[ruled,vlined]{algorithm2e}

\newcommand{\norm}[1]{\left\lVert#1\right\rVert}

\usepackage{colortbl}%
\makeatletter

\newcommand{\columnname}[1]
{\makebox[\tempwidth][c]{#1}}

\usepackage{graphics} 
\usepackage{graphicx}
\usepackage{amsmath} 

\usepackage[pagebackref=true,breaklinks=true,colorlinks,bookmarks=false]{hyperref}

\def\cvprPaperID{****} 
\def\confYear{CVPR 2021}

\newcommand\blfootnote[1]{%
	\begingroup
	\renewcommand\thefootnote{}\footnote{#1}%
	\addtocounter{footnote}{-1}%
	\endgroup
}

\begin{document}

\title{Attention-based Adversarial Appearance Learning of Augmented Pedestrians}

\author{Kevin Strauss$^{1*}$ and Artem Savkin$^{1,2*}$ and Federico Tombari$^{1,3}$
\vspace{0.1cm}
\\
$^{1}$ Technical University of Munich \hspace{0.3cm}
$^{2}$ BMW  \hspace{0.3cm}
$^{3}$ Google
}

\maketitle

\blfootnote{*Equal contribution}

\begin{abstract}

Synthetic data became already an essential component of machine learning-based perception in the field of autonomous driving. Yet it still cannot replace real data completely due to the sim2real domain shift. In this work, we propose a method that leverages the advantages of the augmentation process and adversarial training to synthesize realistic data for the pedestrian recognition task. Our approach utilizes an attention mechanism driven by an adversarial loss to learn domain discrepancies and improve sim2real adaptation. Our experiments confirm that the proposed adaptation method is robust to such discrepancies and reveals both visual realism and semantic consistency. Furthermore, we evaluate our data generation pipeline on the task of pedestrian recognition and demonstrate that generated data resemble properties of the real domain. 
\end{abstract}

\section{Introduction}

Self-driving vehicles envision a significantly beneficial impact on many aspects of the economy and society: higher safety on the road, less time committed to driving, accessibility for a wider range of users, and reduced environmental effect of transportation to name a few. For autonomous driving to evolve from the research area to the application domain though, it is critical for autonomous systems to fulfill the high safety requirements and meet reliability expectations.

Machine learning-based perception components rely heavily on the availability of high-quality large-scale datasets. Those perception models expect the training data to cover similar environments as the deployment or testing data. This is by no means a trivial task to meet such expectations. We are usually not able to identify all use cases in advance and certain scenes do rarely occur and are therefore hard to capture or they may not be possible to recreate due to ethical reasons. To reconstruct a near-accident scenario such as in \cite{Bloomberg20} one would need to put vulnerable traffic users at risk. Also, it is usually laborious and expensive to obtain large amounts of annotated data as labeling requires considerable time-consuming human effort \cite{Cordts2016}.

An attractive approach to responding to the described challenges is synthetically generated data. This idea has already been successfully applied in the research community \cite{Broggi2005}. Synthesizing data is a very cost-efficient process that helps to generate annotated data and enables a controlled generation of traffic scenes. 

However, for real-world applications, their use is relatively limited. According to \cite{Richter2016}, naively training a segmentation model on synthetic data results in degrading prediction accuracy when evaluated on real data. The root cause for this phenomenon can be identified as the domain gap that is present between the synthetic and real domains. It arises from a difference in content and appearance distributions between domains. In literature, this discrepancy is more commonly addressed as the covariate (or domain) shift \cite{Sugiyama2012}.

\begin{figure}[!t]

\begin{minipage}{0.325\columnwidth}
\includegraphics[width=1\textwidth]{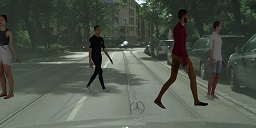}
\includegraphics[width=1\textwidth]{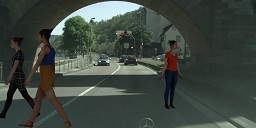}
\end{minipage}
\begin{minipage}{0.325\columnwidth}
\includegraphics[width=1\textwidth]{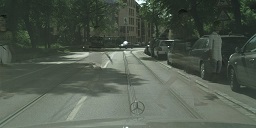}
\includegraphics[width=1\textwidth]{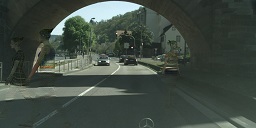}
\end{minipage}
\begin{minipage}{0.325\columnwidth}
\includegraphics[width=1\textwidth]{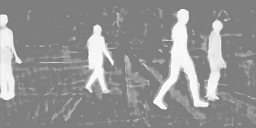}
\includegraphics[width=1\textwidth]{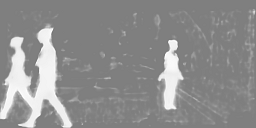}
\end{minipage}

\caption{Examples of image augmented with pedestrian, discrepancy-induced inconsistent adaptation and attention maps learned based on such discrepancy (left to right).}
\label{fig:vanishing_pedestrians}

\end{figure}

The main contribution of this work is a novel two-staged method for unsupervised learning of the realistic appearance of virtual pedestrians that visually align with the realistic surrounding environment.
Our method's intuition resides on the insight that the image regions vanishing throughout the adaptation procedure bear the highest discrepancy between domains (w.r.t style/content). Inspired by \cite{Mejjati2018} we suggest detecting those discrepancy regions employing the attention mechanism attached to adversarial loss in the first step. Predicted discrepancy areas are then utilized in the second step of the multi-discriminator adaptation setup.
We employ a data augmentation pipeline based on \cite{Savkin2020} that blends virtual pedestrians into existing real image scenes to enable a controlled generation of yet unseen urban traffic scenarios involving pedestrians.

We argue that proposed appearance learning is robust against distribution discrepancies between real and synthetic data and thus preserves semantic consistency. In our experiments, we demonstrate that data generated using this method exhibit visual realism and no content perturbations. We additionally evaluate the proposed data generation technique on the downstream task of pedestrian recognition and confirm that synthesized scenes resemble properties of real data.

\section{Related Work}

To enable transfer learning from the synthetic domain to the real many researchers aim to reduce the covariate shift across domains. Most recent works on fully synthetic data generation propose to tackle the shift via photo-realistic rendering, such methods include \textit{PfD} \cite{Richter2016}, \textit{Synthia} \cite{Ros2016} and \textit{Synscapes} \cite{Wrenninge2018}.

Semi-synthetic approaches can employ straight-forward "cut and paste" methods like \cite{Busto2015, Dwibedi2017, Huang2017} or more sophisticated techniques which focus on the augmentation of the static real scenes with CAD objects such as cars \cite{Alhaija2017} or pedestrians \cite{Cheung2016}.

Another research branch employs generative networks based on adversarial training that learn a realistic style of reference scenes and apply it to fully synthetic ones \cite{Shrivastava2016, Zhu2017, Chang2019}. These approaches find the application not only in the self-driving area but also for indoor scene generation \cite{McCormac2017}. Researchers aim to eliminate the aforementioned domain gap between the synthetic and real domain by learning a synthetic-to-real image translation function \cite{Huang2018, Shu2018, Chang2019}. This mapping function is typically approximated by a neural network based on the generative adversarial framework (GAN) \cite{Goodfellow2014}. In such an adversarial training setup two networks, called generator and discriminator, are involved in a \textit{zero-sum game}. Discriminator learns to distinguish reference samples from the ones produced by the generator. The generator in turn learns to counteract. This game settles at a so-called \textit{Nash equilibrium}. By design GAN minimizes the distance between generated and target probability density functions \cite{Goodfellow2014}. GANs indeed achieve visually convincing results but also tend to integrate visual artifacts into generated samples that induce semantic mismatches.

Adversarially induced mismatches could be observed in figure~\ref{fig:vanishing_pedestrians}. To mitigate such an effect, several works propose to integrate constraints to counteract this unwanted behavior \cite{Hoffman2017}. Some works utilize generative networks for augmentation \cite{Ouyang2018, JieWu2019, Vobecky2019} to extend the variation of pedestrian instances. In practice, various GAN variants \cite{Zhu2017, Hoffman2017, Huang2018} are proven capable of providing a set of samples whose distribution is similar to the target data. This could be interpreted as similarity in content and style. However, the source and corresponding generated samples do typically exhibit mismatches in content. In \cite{Hoffman2017} it is argued that aligning marginal distributions does not enforce semantic consistency. These inconsistencies may arise due to the discriminator's ability to incorporate content information into its decision-making process. This drives the generator to perform perturbations in the source images to level out the discrepancies between domains. Elements that are rather uncommon in the target dataset get consequently replaced by others that are more prevalent to minimize the class distribution discrepancy. In other words, the adversarial training dynamics encourage content-modifying translations in the presence of a significant covariate shift.

\section{Approach}

\begin{table}
\centering
\begin{tabular}{l|cc}
    \toprule
    \multicolumn{1}{c}{} &
    \multicolumn{2}{c}{Datasets} \\
    \cmidrule(lr){2-3}
    \multicolumn{1}{l}{Class} &
    \multicolumn{1}{c}{Cityscapes} &
    \multicolumn{1}{c}{Augmented} \\
    \cmidrule(l){1-3}
    \rowcolor[gray]{0.99}
    person & 1.08 & 8.00 \\ 
    road & 32.62 & 30.04 \\ 
    building & 20.21 & 19.09 \\ 
    car & 6.19 & 5.33 \\ 
    \bottomrule
\end{tabular}
\caption {Class balance before and after augmentation. Class \textit{pedestrian} increases by factor 8.}
\label{tab:class_balance}
\end{table}

\subsection{Attention with Adversarial Loss}
\label{sec:semantically_inconsistent_da}

Our setup is based on \cite{Savkin2020}, which augments the Cityscapes dataset \cite{Cordts2016} with 3D pedestrian models and adversarially learns to cast realistic style on those models by means of a \textit{multi-discriminator}. Due to the in-painting of virtual model instances, the pedestrian class becomes the most imbalanced class between the original and augmented dataset (see~\ref{tab:class_balance}). As argued before, the adversarial training objective guides the generator to remove augmented pedestrians to restore the initial real (target) distribution. This has an undesired effect that in-painted pedestrians vanish instead of attaining a realistic look, see figure~\ref{fig:vanishing_pedestrians}. Our approach leverages the ability of adversarial loss to identify image areas where the most prominent discrepancies between domains occur. We employ an attention mechanism to learn those areas to later utilize them in the multi-discriminator.

\begin{figure}[t]
    \includegraphics[width=1\columnwidth]{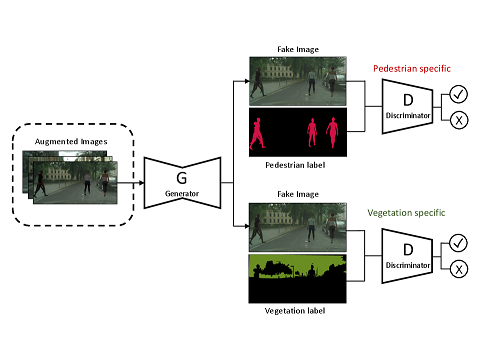}
    \caption{An example of a multi-discriminator network with splits focused on pedestrian and vegetation.}
    \label{fig:multi_discriminator}
\end{figure}

\subsection{Multi-discriminator Architecture}
\label{sec:multi_discriminator_architecture}

As shown in figure~\ref{fig:vanishing_pedestrians}, domain adaptation frameworks based on adversarial training approximate translation functions which implicitly learn to minimize the distribution discrepancy between the source and the target domains. Such functions while translating images from one domain to another may not only modify the style of an image but also its content. In our specific domain adaptation setup with augmented pedestrians, this phenomenon can be observed as in-painted pedestrians disappear during translation. 

To counteract such undesired behavior, several works proposed to split discriminator of the adversarial network into multiple ones to overcome distribution discrepancies \cite{Peilun2018, Savkin2020}. The intuition behind it is to restrict the decisive context of the discriminator and let it consider only specific aspects (e.g. semantic class). The proposed multi-discriminator adversarial network extends the original CycleGAN \cite{Goodfellow2014} framework with additional class-specific discriminators. Each discriminator assesses merely those parts of the input image that belong to the dedicated semantic class. By design, their decision-making freedom regarding the content is eliminated as they are not aware of the class shift existing across domains. Thus, such discriminators only focus on the appearance features of the particular class. The generator is then driven to perform perturbations solely on such appearance features of that class.

This is enabled by splitting the holistic input image into disjoint patches, where each patch corresponds to one particular semantic class. Subsequently, these patches are provided as an input to their corresponding class-specific discriminator. Visual representation of the multi-discriminator is depicted in the figure~\ref{fig:multi_discriminator}.

\begin{figure}[t]
\includegraphics[width=0.45\textwidth]{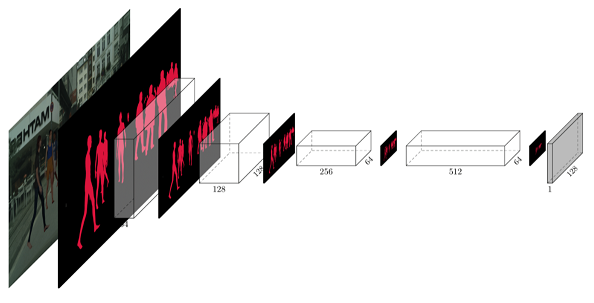}
\caption{Discriminator with its \textit{MaskLayer} introduced after each convolution block.}
\label{fig:masklayer}
\end{figure}


In our setup we denote the source image space as $S$ (augmented pedestrian domain) and the target image space as $T$ (real domain).
Random variables $s$ and $t$ defined in the spaces $S$ and $T$ are independent and identically distributed according to $P_S$ and $P_T$ respectively. They take values $s_i$ and $t_i$ from $ \mathbb{N}^{3 \times h \times w}$, here $h\times w$ is the size of particular image sample. Subsets $\mathcal{X}_{S}$ and $\mathcal{X}_{T}$ of sizes $N_S$ and $N_{T}$ respectively comprise a set of such samples. We denote the corresponding semantic label distribution spaces as $\bar{S}$ and $\bar{T}$, respectively. We apply analogous notations for the labels spaces of source and target domains: 

\begin{align}
    \begin{split}
    &s_i \in \mathcal{X}_{S} \subset S \in \mathbb{N}^{3 \times h \times w},\ i \in [N_S]\\
    &\bar{s}_i \in \mathcal{X}_{\bar{S}} \subset \bar{S} \in \mathbb{N}^{1 \times h \times w},\ i \in [N_{S}]\\
    &(s_i,\bar{s}_i) \sim P_S\\
    \end{split}
\label{eq:definition_source}
\end{align}

\begin{align}
    \begin{split}
    &t_j \in \mathcal{X}_{T} \subset T \in \mathbb{N}^{3 \times h \times w},\ j \in [N_T] \\
    &\bar{t}_j \in \mathcal{X}_{\bar{T}} \subset \bar{T} \in \mathbb{N}^{1 \times h \times w},\ j \in [N_{T}]\\
    &(t,\bar{t}) \sim P_T\\
    \end{split}
\label{eq:definition_target}
\end{align}

We aim to estimate a mapping function $G_{S}: S \rightarrow T$  based on $\mathcal{X}_{S}$ and $\mathcal{X}_{T}$. Similar to the baseline CycleGAN \cite{Zhu2017} approach we additionally estimate $G_{T}: T \rightarrow S$ in order to enable the cyclic consistency constraint. Estimation procedure is called supervised if involves $\mathcal{X}_{\bar{S}}$ and $\mathcal{X}_{\bar{T}}$, otherwise unsupervised.

In the baseline approach discriminators $D_S$ and $D_T$, where $D_S$ aims to distinguish between source images $s \in S$ and translated target images $G_{T}(t), t \in T$; in the same way, $D_T$ aims to discriminate between $t \in T$ and $G_{S} (s),  s \in S $.Contrary to that we introduce class-specific discriminators for any particular class $c$ and denote them as $D^c_S$ and $D^c_T$. We employ PatchGAN descriminator \cite{Isola2016} architecture which maps provided image to a $n \times n$ vector: $D(s) \in \mathbb{R}^{n \times n}$

In order to split an input image into disjoint class-patches mentioned in section~\ref{sec:multi_discriminator_architecture}, we apply a binary mask $M^c(\bar{s}) $ and its down-sampled versions after each (a) convolutional layer of the discriminator and (b) output layer. The mask $M^c(\bar{s}) \in \{0,1\}^{h \times w}$ is obtained from the semantic map $\bar{s}$ and indicates those positions from the corresponding source image $s$ that are represented by class $c$. (a) is manifested by adding the binary mask to the input parameters of a specialized discriminator as $D^c_T (t,M^c(\bar{t}))$, and (b) is enforced by introducing an operation $\odot$ that applies a resized binary mask to the output of the discriminator. For (a) and (b) $M^c(\bar{s}) $ is down-sampled to match the dimensions of the feature maps and output layer. This functionality is enabled by  \textit{MaskLayer} which is shown in figure~\ref{fig:masklayer} \cite{Savkin2020}.


Inspired by the objective of adversarial loss in LSGAN \cite{Mao2017} our class-specific adversarial objective looks as follows:

\begin{align}
    \begin{split}
    L^S_{adv}
    &(G_{S}, D^c_T)=\\
    &\mathbb{E}_{(t,\bar{t}) \sim p_T} \left[\norm{(D^c_T(t,M^c(\bar{t})) - \mathds{1}) \odot M^c(\bar{t})}_F^2 \right] +\\
    &\mathbb{E}_{(s,\bar{s}) \sim p_S} \left[\norm{D^c_T(G_{S}(s),M^c(\bar{s})) \odot M^c(\bar{s})}_F^2 \right]
    \label{eq:masked_loss}
    \end{split}
\end{align}

Here $\mathds{1} \in \mathbb{N}^{n \times n}$ denotes a matrix of ones, and $\norm{\cdot}_F$ denotes the Frobenius norm. It is worth mentioning that in case of $M^c(\bar{s}) = M^c(\bar{t}) = \mathds{1}$, where we eliminate any masking, we do not reduce the awareness of the split discriminator at all and retrieve the original adversarial loss objective as specified in \cite{Mao2017}. Similarly we define $L^T_{adv}(G_{T}, D^c_S)$.

\subsection{Annotation-based Split}\label{sec:annotation_based_split}

For practical scenarios, the question arises which splitting strategy should be devised. This question may require task-specific adaptation.
In our vanishing pedestrian problem, for instance, the pedestrian class was exposed to major semantic mismatches while the background has been left almost unchanged (figure~\ref{fig:vanishing_pedestrians}). That is, in accordance with our observations, classes which exhibit significant discrepancies are typically more prone to semantic mismatches.

Based on this observation, an effective splitting strategy for a two-discriminator setup could be the following: one discriminator is assigned to the class with the largest difference in apparition frequency across domains - \textit{pedestrian}, while the other focuses on the background.
We denote the pedestrian and background class in the following as $p$ and $b$, respectively. The overall objective takes now an aggregated form with class-specific components in both domains: 
\begin{align}
\begin{split}
L & (G_{S}, G_{T}, D^p_T, D^b_T, D^p_S, D^b_S ) =\\
& \underbrace{L^S_{adv}(G_{S}, D^p_T) + L^S_{adv}(G_{S }, D^b_T)}_{\text{Source Domain Split}} \ +\\
& \underbrace{L^T_{adv}(G_{T}, D^p_S ) + L^T_{adv}(G_{T}, D^b_S )}_{\text{Target Domain Split}}  \ + \\
&\lambda_{cyc}(L^S_{cyc} (G_{S},G_{T}) +L^T_{cyc} (G_{S},G_{T}))
\label{eq:fullobjective}
\end{split}
\end{align}

Here, $L_{cyc}^d$ represents the cyclic consistency loss that enforces the reconstruction of an image in the domain $d$ weighted by $\lambda_{cyc}$ to control the relative importance between cyclic and adversarial objectives. In the main part of the algorithm~\ref{algo:training} (step 2) we aim to solve:
\begin{align}
\begin{split}
\min\limits_{G_{S},G_{T}} \max\limits_{D^p_T, D^b_T, D^p_S, D^b_S} L(G_{S}, G_{T}, D^p_T, D^b_T, D^p_S, D^b_S) 
\label{eq:optimisation}
\end{split}
\end{align}

\begin{figure}[t]
\includegraphics[width=1\columnwidth]{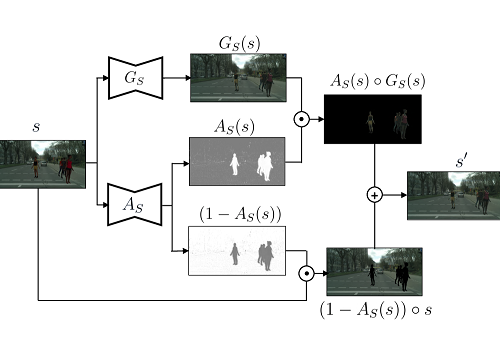}
\caption{Flow diagram from the source domain $S$ to the target domain $T$.}
\label{fig:attprocess}
\end{figure}

\begin{algorithm}[t]
\SetAlgoLined
\KwResult{ $G_S$ that enables realistic translation of augmented pedestrians in a semantic consistent manner }
initialize $G_S, A_S, D_T$ and $G_T, A_T, D_S$\;
\For{training iterations}{
    sample \textit{m}-minibatch ${s_1,...,s_m}$ from $X_S$\;
    sample \textit{m}-minibatch ${t_1,...,t_m}$ from $X_T$\;
    update $G_S, A_S, D_T$ and $G_T, A_T, D_S$ using \ref{eq:vanlosses}
}
derive $M^S(s)$ and $M^T(t)$ $\forall s \in X_S$ and $ \forall t \in X_T$ from $A_S(s)$ and $A_T(t)$  \;
reinitialize $G_S, D_T$ and $G_S, D_T$\;
\For{training iterations}{
    sample \textit{m}-minibatch ${s_1,...,s_m}$ from $X_S$\;
    sample \textit{m}-minibatch ${t_1,...,t_m}$ from $X_T$\;
    update $G_S, D^p_T$ and $G_T, D^p_S$ using \ref{eq:fullobjective} and using masks $M^S(s)$ and $M^T(t)$
}
\caption{Training algorithm}
\label{algo:training}
\end{algorithm}

\subsection{Attention-guided Split}\label{sec:attention_guided_split}

A multi-discriminator network from~\ref{sec:annotation_based_split} is, therefore, a supervised model in the sense that it requires ground truth annotation to learn a semantically consistent translation mapping. This requires manual annotation and involves a handcrafted masking policy. To remedy these limitations we extend our model so that it can find an adequate splitting policy by learning the split regions.

The key idea is to leverage the ability of the discriminator to reveal a discrepancy between source and target domains. We intend to locate those areas of the image which provide a signal for the adversarial training. Those areas are then used as masks in a multi-discriminator setup to restrict where the actual translation occurs.

For this purpose, inspired by the works of \cite{Chen2018, Mejjati2018} we devise an attention mechanism on top of the multi-discriminator architecture. Attention maps are learned and shaped throughout the training process by the same adversarial signal that enables the discriminator to distinguish source and target domains. The regions which are characteristic of a domain are therefore most likely to be included in the attention map. One can think of the attention mechanism driven by adversarial loss as a magnifying glass for those pixels which reveal a major discrepancy between domains.

The framework we use to generate attention maps is based on the multi-discriminator model from~\ref{sec:annotation_based_split} extended with two additional attention networks $A_S$ and $A_T$ as in \cite{Mejjati2018}. We denote $A_S: S \to S_A$ and $A_T: T \to T_A$, where $S_A$ and $T_A$ are the domains of attention maps induced from $S$ and $T$, respectively. The attention maps have continuous values in $[0,1]$ and the same size as the images in the source and the target domain so that they can be used to constrain the output of the generator to relevant image regions.
The application of attention masks is by using the element-wise Hadamard product $\circ$ between the generated image and the attention map.

\begin{align}
s'=\underbrace{A_S(s) \circ G_{S}(s)}_{\text{Translated Image Region}} + \underbrace{(1-A_S(s)) \circ s}_{\text{Unchanged Image Region}} 
\label{eq:attentiontranslation}
\end{align}

The process of applying the attention is visualized in figure~\ref{fig:attprocess}.

%
To learn the attention maps, we employ a single discriminator on the whole image to achieve attention learning.
Due to the attention-adapted generation process, the adversarial energy is now given by:
\begin{align}
\begin{split}
L^{S}_{adv}
&(G_{S}, A_S, D^c_T ) =\\  
& \mathbb{E}_{t \sim p_T(t)} \left[ \frac{1}{n^2} \norm{D^c_T (t,M^a) \odot M^a}_F^2 \right] + \\ 
& \mathbb{E}_{s \sim p_S(s)} \left[ \frac{1}{n^2} \norm{(D^c_T (\smash{s',M^a})  - \mathds{1}) \odot M^a}_F^2 \right]
\end{split}
\label{eq:adv2}
\end{align}

To stay consistent with the formulation in algorithm \ref{algo:training} we accomplish attention-learning in a two-discriminator setup by setting one mask as all ones and the other as all zeros. This way, we have effectively only one single discriminator that is aware of the image distribution while the other does not contribute to learning. For this purpose we use two masks: $M^a=\mathds{1}$ and $M^{\emptyset}=O$ where $a$, $\emptyset$ and $O$ denote "all classes", "no classes" and the zero matrix respectively. This way, we have effectively only one single discriminator that is aware of the entire image distribution. 

The final aggregated loss of step 1 of the algorithm~\ref{algo:training} is given by the sum of adversarial and cyclic losses for the source and target domain:
\begin{align}
\begin{split}
L(G_{S},&G_{T},A_S,A_T, D^a_T, D^{a}_S) =\\
&L^S_{adv}(G_{S},A_S, D^a_T ) + L^T_{adv}(G_{T},A_T, D^a_S)\\
&\lambda_{cyc} (L^S_{cyc}+L^T_{cyc})
\end{split}
\label{eq:vanlosses}
\end{align}

Finally, we formulate the complete training algorithm~\ref{algo:training}.

\subsection{Intermediate Translation Domain}

We apply our attention-based multi-discriminator framework using an attention-based split to learn a domain adaptation mapping from synthetic pedestrians \textit{Augmented} to real pedestrians \textit{Cityscapes}.
The sub-distributions induced by the attended regions in both domains can exhibit discrepancies with regard to content. However, the split discriminator should be exposed via the attention maps to the same content.

\begin{figure}[t]

\begin{minipage}{0.325\columnwidth}
\includegraphics[width=1\textwidth]{images/nopedaug2noped/epoch096_real_A}
\includegraphics[width=1\textwidth]{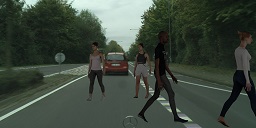}
\includegraphics[width=1\textwidth]{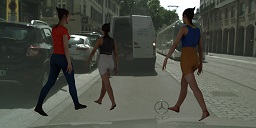}
\includegraphics[width=1\textwidth]{images/nopedaug2noped/epoch110_real_A}
\includegraphics[width=1\textwidth]{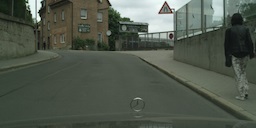}
\includegraphics[width=1\textwidth]{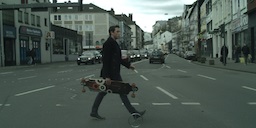}
\includegraphics[width=1\textwidth]{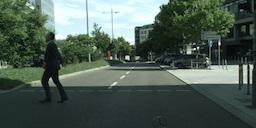}
\includegraphics[width=1\textwidth]{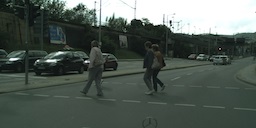}
\end{minipage}
\begin{minipage}{0.325\columnwidth}
\includegraphics[width=1\textwidth]{images/nopedaug2noped/epoch096_fake_B}
\includegraphics[width=1\textwidth]{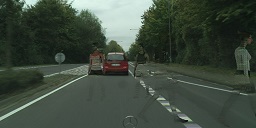}
\includegraphics[width=1\textwidth]{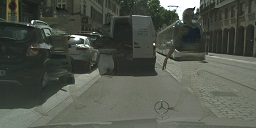}
\includegraphics[width=1\textwidth]{images/nopedaug2noped/epoch110_fake_B}
\includegraphics[width=1\textwidth]{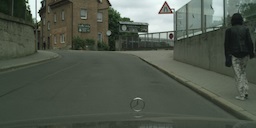}
\includegraphics[width=1\textwidth]{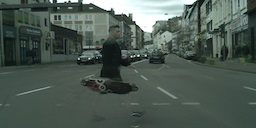}
\includegraphics[width=1\textwidth]{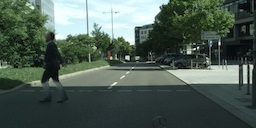}
\includegraphics[width=1\textwidth]{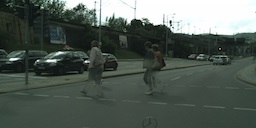}
\end{minipage}
\begin{minipage}{0.325\columnwidth}
\includegraphics[width=1\textwidth]{images/nopedaug2noped/epoch096_att_A}
\includegraphics[width=1\textwidth]{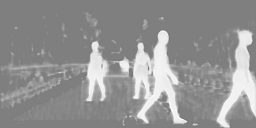}
\includegraphics[width=1\textwidth]{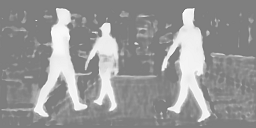}
\includegraphics[width=1\textwidth]{images/nopedaug2noped/epoch110_att_A}
\includegraphics[width=1\textwidth]{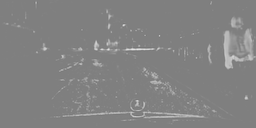}
\includegraphics[width=1\textwidth]{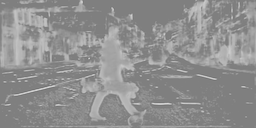}
\includegraphics[width=1\textwidth]{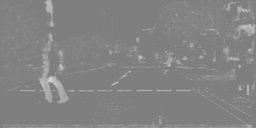}
\includegraphics[width=1\textwidth]{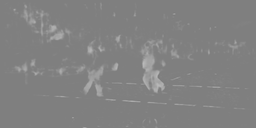}
\end{minipage}

\caption{Examples of the attention maps produced during the training for the translation into intermediate domain (left to right: original, adapted, attention).}
\label{fig:attention_nopedaug2noped}
\end{figure}

In one translation direction (\textit{Augmented} to \textit{Cityscapes}) clear attention regions limited only to pedestrian content evolve while in the other direction almost the whole image is highlighted by attended regions, see top half rows of figure \ref{fig:attention_nopedaug2noped}.
We generalize our findings and attribute them to the case when an underrepresented class (real pedestrians) is shifted to an over-represented class (virtual pedestrians), which enables attention maps learning in both directions.

\begin{figure}[b]
\includegraphics[width=1\columnwidth]{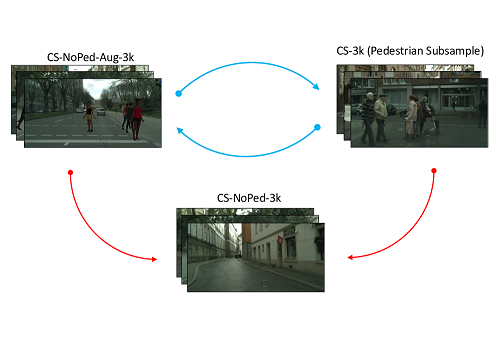}
\caption{Instead of traditional source-target adaptations (blue arrows), attention maps are learned via adaptations to an intermediate translation domain (red arrows)}\label{fig:intermediate_domain}
\end{figure}

We, therefore, propose a novel strategy in which attention regions are now learned through domain adaptation mappings from the synthetic and real to an intermediate domain. A schema of this approach is displayed in figure \ref{fig:intermediate_domain}. The intermediate domain represented by \textit{No pedestrians} is characterized by having no pedestrian information at all. By doing so the pedestrian class always shifts to an underrepresented mode (no pedestrian). The intuition behind this strategy is that discrepancy between intermediate domain and domains of interest is most prominent so it intensifies learning of the attention maps.

By visual inspection of the last two rows of figure \ref{fig:attention_nopedaug2noped} we can confirm that real pedestrians tend to get rendered away and thus the attention maps are guided to partly focus on pedestrian-specific regions. 

\begin{figure*}[t!]

\begin{minipage}{0.195\textwidth}
\includegraphics[width=1\textwidth]{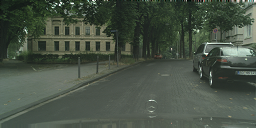}
\includegraphics[width=1\textwidth]{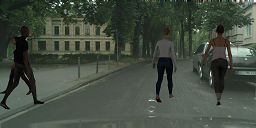}
\includegraphics[width=1\textwidth]{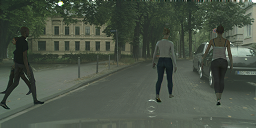}


\end{minipage}
\begin{minipage}{0.195\textwidth}
\includegraphics[width=1\textwidth]{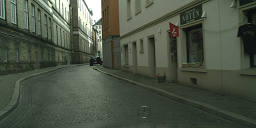}
\includegraphics[width=1\textwidth]{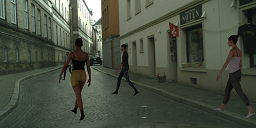}
\includegraphics[width=1\textwidth]{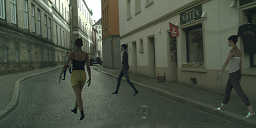}


\end{minipage}
\begin{minipage}{0.195\textwidth}
\includegraphics[width=1\textwidth]{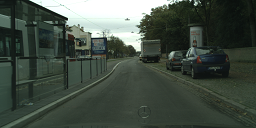}
\includegraphics[width=1\textwidth]{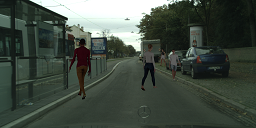}
\includegraphics[width=1\textwidth]{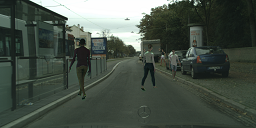}

\end{minipage}
\begin{minipage}{0.195\textwidth}
\includegraphics[width=1\textwidth]{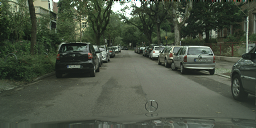}
\includegraphics[width=1\textwidth]{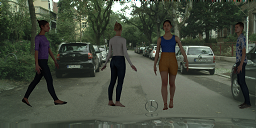}
\includegraphics[width=1\textwidth]{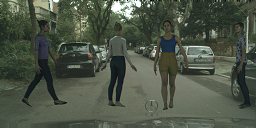}
\end{minipage}
\begin{minipage}{0.195\textwidth}
\includegraphics[width=1\textwidth]{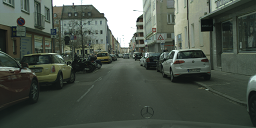}
\includegraphics[width=1\textwidth]{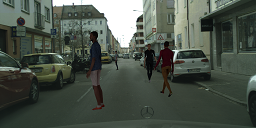}
\includegraphics[width=1\textwidth]{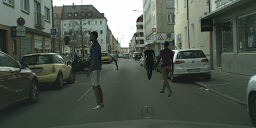}
\end{minipage}

\caption{Sample images from \textit{No Pedestrians} (top) together with their \textit{Augmented} (middle) and \textit{Adapted} (bottom) counterparts.}
\label{fig:results_transfer}
\end{figure*}

\section{Evaluation}

\subsection{Datasets}

In this section, we aim to demonstrate that our approach enables data synthesis with appearance characteristics that reproduce real data concerning pedestrian elements. To do so we created a novel non-pedestrian dataset, apply our augmentation with an appearance learning approach and analyze to which extent the pedestrian distribution of the generated samples resembles representative samples of the real world.
We focused on the Cityscapes \cite{Cordts2016}, a large-scale dataset of complex urban traffic scenes recorded in different cities across Germany. Cityscapes provides ~3k images with \textit{fine} and ~20k with \textit{coarse} pixel-dense annotations. They include semantic and instance labels as well as disparity maps. The latter is decisive for picking this particular dataset. Stereo pairs provide spatial information about the traffic scene which is crucial for the augmentation process.

We aim to design an experiment in which the effect of our approach becomes evident. For this purpose, we created a customized dataset that does not contain any pedestrian information. This could be done by leveraging instance segmentation labels from the original training data. We acquire 646 samples from \textit{fine} and 2329 samples from \textit{coarse} subsets of Cityscapes which do not contain pedestrian instances (ratio of pixels is $<0.05$ for \textit{fine} and $=0$ for \textit{coarse}). As a result, our customized datasets contain 2975 samples in total and equals \textit{train} in size. We denote further our custom dataset as \textit{No pedestrians}.

In the final step of our experimental setup, we applied the augmentation strategy from \cite{Savkin2020} to enrich our custom dataset with pedestrian instances. This strategy picks randomly a number of objects from the pool of CAD pedestrians and puts them onto a so-called \textit{spawn map} - set of collision-free spots on the ground. We denote the dataset retrieved in this way as \textit{Augmented}.

\begin{figure*}[ht]

\begin{minipage}{0.195\textwidth}
\includegraphics[width=1\textwidth]{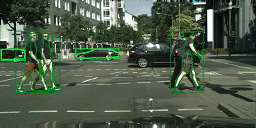}
\vspace{0.05cm}
\includegraphics[width=1\textwidth]{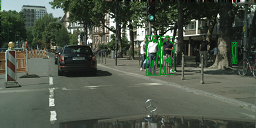}
\end{minipage}
\begin{minipage}{0.195\textwidth}
\includegraphics[width=1\textwidth]{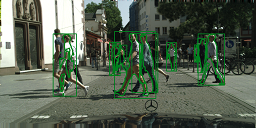}
\includegraphics[width=1\textwidth]{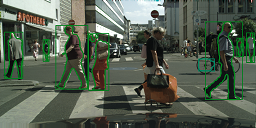}
\end{minipage}
\begin{minipage}{0.195\textwidth}
\includegraphics[width=1\textwidth]{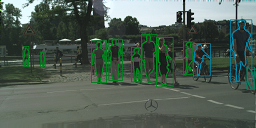}
\includegraphics[width=1\textwidth]{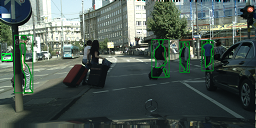}
\end{minipage}
\begin{minipage}{0.195\textwidth}
\includegraphics[width=1\textwidth]{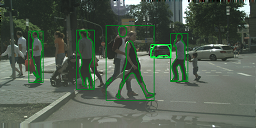}
\includegraphics[width=1\textwidth]{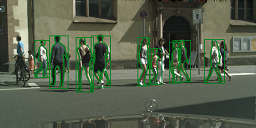}
\end{minipage}
\begin{minipage}{0.195\textwidth}
\includegraphics[width=1\textwidth]{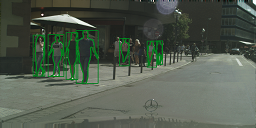}
\includegraphics[width=1\textwidth]{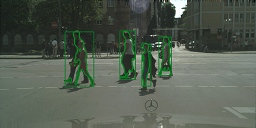}
\end{minipage}

\caption{Pedestrian detection examples provided by the Mask R-CNN trained on the adapted images.}
\label{fig:results_detection}
\end{figure*}

\begin{table*}[t!]
\centering
\footnotesize
\begin{tabular}{lcccccccc}
\toprule
\multicolumn{1}{c}{} &
\multicolumn{2}{c}{Cityscapes \textit{train}} &
\multicolumn{2}{c}{Augmented} &
\multicolumn{2}{c}{Adapted}\\
\cmidrule(lr){2-3}\cmidrule(lr){4-5}\cmidrule(lr){6-7}\cmidrule(lr){8-9}
\multicolumn{1}{l}{} &
\multicolumn{1}{c}{AP} &
\multicolumn{1}{c}{Ratio} &
\multicolumn{1}{c}{AP} &
\multicolumn{1}{c}{Ratio} &
\multicolumn{1}{c}{AP} &
\multicolumn{1}{c}{Ratio}\\
\cmidrule(lr){2-9}
Cityscapes \textit{eval} & 0.360 & 100 \% & 0.092 & 25.56 \% & 0.118 & 32.78 \% \\
CS-Val-60 & 0.480 & 100 \% & 0.192  & 40.00 \% & 0.222  & 46.25 \% \\
CS-Val-80 & 0.512 & 100 \% & 0.247 & 48.24 \% & 0.292 & 57.03 \% \\
CS-Val-Filt-60 & 0.516 & 100 \% & 0.268 & 51.94 \% & 0.288 & 55.81 \% \\
CS-Val-Filt-80 & 0.547 & 100 \% & 0.302 & 55.21 \% & 0.320 & 58.50 \% \\
\bottomrule
\end{tabular}
\caption {Pedestrian detection results (Average Precision) for Mask R-CNN obtained on different train (column) and validation (row) datasets. Additionally, ratio to the upper bound is reported.}
\label{tab:results}
\end{table*}

\subsection{Experiments}\label{sec:experiments}

\textbf{Qualitative Evaluation.}
First, we enhance the \textit{No Pedestrians} dataset described in the section~\ref{sec:experiments} with augmented pedestrian 3D instances. As previously mentioned we denote it as \textit{Augmented}. As a next step, we apply the attention-guided multi-discriminator adversarial network described in \ref{sec:attention_guided_split} to make a target-alike style onto augmented pedestrian objects. A dataset retrieved by the appearance learning framework in this manner we denote \textit{Adapted}

At this stage, we assess the visual quality of generated images of the \textit{Adapted}. The results of this adaptation could be observed in figure \ref{fig:results_transfer}. Here our first criterion is the actual appearance learning. The adapted images in figure \ref{fig:results_transfer} confirm that over the course of domain transfer training, augmented objects acquire a target-alike look. This is mainly confirmed by the color scheme applied to them and also by the light reflections on some body parts resembling the Cityscapes-alike lighting.

Our second goal is to retain the semantic consistency of the inferred images during the domain adaptation. Contrary to the data retrieved by traditional style transfer approaches, augmented objects in our method do not vanish during our domain transfer training.

\textbf{Quantitative Evaluation.}
To estimate the quality of the data generated by our pipeline, we report recognition performance on the downstream task of detection. We train a Mask R-CNN model on each of the aforementioned datasets and consider the instance segmentation performance. We apply our Mask R-CNN model with the ResNet-FPN-50 backbone \cite{Lin2017} and train it with randomly cropped samples of size $800 \times 1024$ pixels. We choose the batch size to be 8 and train the model for 4k iterations following the original training setup of \cite{He2017}.

We evaluate the trained models on Cityscapes \textit{val} set, which comprises 500 images, semantic and instance segmentation labels. We report the AP (Average Precision) metric achieved by the Mask R-CNN model for the pedestrian class in the first line of the table~\ref{tab:results}. The reported values achieved for different training sets could be observed in 4 columns. As one can expect Mask R-CNN trained on the original Cityscapes \textit{train} shows the best result with AP of $0.360$. This serves as the upper bound in our evaluation. Training on the \textit{Adapted} reaches AP $0.118$ which is only around $33\% $ of best performance but almost $10\%$ better as pure augmented data \textit{Augmented} with it's AP $0.092$. The mere augmentation of virtual pedestrians caused a recognizable performance improvement. For the translated datasets \textit{Adapted} the corresponding AP values increased even more. Examples of such segmentation could be observed in figure \ref{fig:results_detection}.

However, we still observe a significant gap with the training on the original dataset. This could indicate that our approach has not been able to reproduce the same variability or visual appearance of pedestrians that are present in the target data. We presume the root cause for that lies in the fact that the augmentation process itself is still relatively limited with respect to multiple aspects such as the variance of pedestrian types or their distance. For instance, we only operate with 40 CAD models. 

\textbf{Ablation Study.}
To study the limitations of our augmentation and appearance-learning models we aim to restrict the validation dataset in such a way that it reflects the pedestrian variance similar to augmentation data. 

We, therefore, restrict the original validation dataset \textit{CS-Val} in two ways: (1) to account for the fact that augmented pedestrians are mostly placed in the foreground of the picture while real pedestrians rather appear in the background we create two novel validation sets which comprise pedestrian instances whose pixel height is exclusively greater than 60 pixels and 80 pixels and denote them as \textit{CS-Val-60} and \textit{CS-Val-80}, respectively, and (2) we manually exclude specific scenes where the detection network typically fails e.g. if pedestrians are severely exposed to occlusion or captured in a non-typical pose (e.g. sitting). As a result we obtain the novel validation datasets: \textit{CS-Val-Filt-60} and \textit{CS-Val-Filt-80}.

In the last bottom table \ref{tab:results} we report AP metric for the Mask R-CNN results on the new validation sets as well. They reveal that the performance is highest (with AP $0.361$ and $67\%$ of upper bound) when the distribution of the evaluation data our augmented data is closest. Based on these results, we can conclude that our augmentation and appearance-learning approach is in fact able to recreate the real distribution that is present in our modified validation sets better. We leave through the study of the effects of the particular augmentation parameters for future work.

\section{Conclusion}

In this work, we demonstrated how adversarial training could introduce semantic inconsistencies during the \textit{sim2real} adaptation. We claim those image regions where such artifacts occur bear the most prominent discrepancies between domains. Our reasoning is, that by design, the discriminator identifies most domain-characteristic features and drives a generator to level them out over the curse of adversarial training.
To learn such discrepancy maps we propose a method that utilizes an attention mechanism driven by the adversarial loss. These attention maps are then integrated into the data generation pipeline as masks for our multi-discriminator architecture.

We also show that this adaptation pipeline with specialized discriminators produces semantically consistent images with augmented pedestrians. Furthermore, it can learn the target-data-alike pedestrian appearance and apply it onto in-rendered CAD models.

To enforce attention maps to highlight regions with aligning contents across domains we introduce an intermediate \textit{non-pedestrian domain}. It indeed improves the alignment of the attention regions.

We performed the semi-supervised sim-to-real translation of augmented pedestrians using our framework and observed that it can reproduce large parts of real-world pedestrian variability.

Although downstream task showed significant improvement compared to pure synthetic data, original same-domain training data remains an upper bound. For that matter, we identified certain statistical aspects which help to replicate original data closer. However, we left a comprehensive study of those aspects for future experiments.

{\small
\bibliographystyle{ieee_fullname}
\bibliography{literature}
}

\end{document}